\documentclass{article}

\usepackage{microtype}
\usepackage{graphicx}
\usepackage{subcaption}
\usepackage{booktabs} 
\usepackage{tabularx}
\usepackage{multirow}
\usepackage{threeparttable}

\usepackage{hyperref}

\usepackage[accepted]{icml2026}

\usepackage{amsmath}
\usepackage{amssymb}
\usepackage{mathtools}
\usepackage{amsthm}

\usepackage[capitalize,noabbrev]{cleveref}

\theoremstyle{plain}

\theoremstyle{definition}

\theoremstyle{remark}

\usepackage[textsize=tiny]{todonotes}

\icmltitlerunning{Decision Titan: Test-Time Training for Long-Term Memory in Offline Reinforcement Learning}

\begin{document}

\twocolumn[
  \icmltitle{Decision Titan: Test-Time Training for\\Long-Term Memory in Offline Reinforcement Learning}



  \icmlsetsymbol{equal}{*}

  \begin{icmlauthorlist}
    \icmlauthor{Jude Waide}{yyy}
    \icmlauthor{Robert Lieck}{yyy}

  \end{icmlauthorlist}

  \icmlaffiliation{yyy}{Department of Computer Science, University of Durham, Durham, United Kingdom}

  \icmlcorrespondingauthor{Jude Waide}{jude.waide@durham.ac.uk}

  \icmlkeywords{Machine Learning, ICML}

  \vskip 0.3in
]



\printAffiliationsAndNotice{}  

\begin{abstract}
   Long-term dependencies remain a major challenge for sequential decision-making in the field of AI: RNNs suffer from vanishing gradients and the limited expressivity of vector-based hidden states, whilst Transformer-based models are limited by the quadratic scaling of attention. Recent work has proposed tackling this problem with the Test-Time Training (TTT) framework, which stores episodic memories in the parameters of a neural network through gradient descent at both train and test-time. This approach has seen success in the domain of Natural Language Processing, however, to the best of our knowledge it has not yet been applied to the domain of Reinforcement Learning (RL), nor has there been a study analysing how this memory practically functions. In this paper, we study the potential of the TTT framework for offline RL by augmenting a Decision Transformer with TTT layers, dubbed the Decision Titan. We analyse performance and properties of the model in the X-Maze environment, an extension of T-Maze designed to test sequential memory, and investigate how the memory mechanism learns by visualising gate values over time. 
   Our key findings are that Decision Titan can learn long-term dependencies with ranges 20x longer than the context window, generalises to lengths 1.7x the training data, but crucially temporal generalisation depends on the time embeddings used, and the ability to learn long-term dependencies depends on how the relevant information is encoded.
\end{abstract}

\section{Introduction}\label{sec:introduction}
Long-term dependencies remain a fundamental challenge across the field of AI. They occur when the optimal output of a neural network depends on information from the distant past---a situation that occurs in many real-world scenarios.
%
Modelling these dependencies is difficult for two main reasons: credit assignment and episodic memory. The credit assignment problem \cite{CAP} occurs when actions have delayed effects on rewards, e.g. an optimal move in a game of chess may only lead to an enemy piece getting captured several moves later. In such a scenario, a model must be able to learn which past actions are responsible for present outcomes---a challenge that becomes increasingly severe over long temporal horizons \cite{LongtermDependencies}. However, in the context of long-term dependencies, the problem shifts: rather than asking \textit{“Which actions are relevant to the current reward?”}, we instead ask \textit{“Which observations are relevant to the current decision?”}. Addressing this form of credit assignment is essential for identifying what information must be retained over time. Once identified, the model must retain it, motivating the need for an effective episodic memory mechanism. 

Current approaches are limited at handling these dependencies. Recurrent Neural Networks (RNNs) \cite{LSTM, GRU} propagate a hidden state through time, theoretically allowing the modelling of arbitrarily long dependencies. However, in practice they suffer from finite hidden-state expressivity and vanishing gradients over time. Transformers \cite{Transformers} avoid these issues by feeding the model the last $k$ inputs at each step, directly modelling the dependencies within this fixed-size window through self-attention. The major issue is the quadratic scaling cost of attention, limiting the size of the context-window and the length of dependencies that can be modelled, positioning Transformers as a \emph{short-term} memory. Whilst there has been work reducing the scaling to linear 
\cite{EfficientTransformers}, the context length is still limited and therefore does not address the underlying need for an episodic \textit{long-term} memory.  

Recent work has proposed storing episodic memories in weights via the Test-Time Training framework \cite{TTT}. By reserving a fraction of a model's parameters as \emph{fast weights}, episodic information can be stored in these weights by using gradient descent at test-time. The Titans architecture \cite{Titans} has applied this framework to Transformers, using self-attention as a short-term memory and TTT layers as a long-term one. Furthermore, they add momentum and a forgetting gate to the fast weight gradient descent, achieving SOTA performance in NLP tasks featuring long-term dependencies. However, to the best of our knowledge, neither Titans nor the TTT framework has been applied to Reinforcement Learning, nor has an in-depth analysis been done on how Titans learn in practice.



In this work, we investigate the potential of the Test-Time Training framework to address long-term dependencies in Reinforcement Learning (RL). To this end, we examine whether Titans can learn an effective policy in environments with long-term dependencies and provide similar benefits in RL as observed in NLP. To answer this, we introduce the Decision Titan (DTi) by merging Titans with Decision Transformer \cite{DecisionTransformer} (a method for applying Transformers to offline RL by treating the problem as a sequence modelling task). 

Our secondary objective is to better understand how the Titans memory module operates and to characterise its strengths and limitations. To achieve this, we introduce X-Maze, an extension of T-Maze \cite{CreditAssignmentVsLongTermDep}, providing a controlled environment for evaluating a model’s ability to remember the \emph{order} of past observations over long temporal horizons. We test DTi in this environment, examining properties such as the ability of the model to extrapolate and the impact of different positional embedding strategies. Furthermore, we visualise the gating mechanisms of the fast weight update rule and perform ablation studies on the momentum. We then evaluate our findings to highlight promising directions of future study. 


We find that the Decision Titan can learn dependencies with ranges 20x longer than the context window. Furthermore, we find that momentum helps the model learn dependencies by propagating them further during training.
We also show that: DTi memory learns which information to remember by chance, catastrophic forgetting is likely the largest limitation facing the approach, the model can remember sequential information under the right conditions, the choice of embedding has an impact on the memory retention of the model when extrapolating to longer episode lengths, and that sparse memory access is a promising area for future research.


\section{Related Work}  \label{sec:related_work}

Our work is not the first to build on the Test-Time Training framework or to extend the Decision Transformer. Some extensions focus on long-term memory but differ in their choice of storage medium, whose theoretical implications we discuss in Section \ref{sec:evaluation}. Here, we briefly review these directions.

\subsection{Test-Time Training Developments}

A limitation of the TTT framework is the quality-efficiency trade-off when selecting the inner-loop chunk size \cite{TTT}, described in Appendix \ref{sec:methodology_background_ttt}. \citeauthor{LaCT} (\citeyear{LaCT}) observe that this limitation applies only when intra-chunk order is important. By instead treating each chunk as an unordered set and scaling chunk sizes up to one million tokens, they achieve far greater GPU utilisation in video-processing tasks. This approach is ill-suited to our setting, where remembering the order of intra-chunk information is central to the task, however work looking at ways to make the TTT framework more efficient is essential for its wider adoption.

An interesting area of study is the choice of architecture for the memory module. Motivated by the linear computational cost of scaling MLP memory capacity, \citeauthor{FastWeightProductKeyMemory} (\citeyear{FastWeightProductKeyMemory}) utilise product-key memory \cite{ProductKeyMemory} instead. This results in a sparse memory access scheme, reducing the complexity to the square root of the number of storage slots. We utilise MLPs in this work as the tasks we are interested in do not require large memory capacity, however our results suggest that such sparse access schemes would improve the model's ability to extrapolate, marking a promising direction of future study. We discuss this further in Section \ref{sec:evaluation}. 

\subsection{Decision Transformer Extensions}

\citeauthor{ODT} (\citeyear{ODT}) utilise a stochastic policy and a special replay buffer, allowing DT to be fine-tuned through online exploration; \citeauthor{PlanningTransformer} (\citeyear{PlanningTransformer}) improve results compared to base DT by conditioning actions on the model's expectation of the future, represented by tokens that are periodically generated by a planning head; \citeauthor{MGDT} (\citeyear{MGDT}) scale DT to play the entire Atari suite \cite{Atari} with a single set of weights, demonstrating that Transformers exhibit the same scaling benefits in RL as in NLP. However, these techniques do not tackle long-term memory, although we note that they are orthogonal to our own approach. Combining them would be an interesting direction of future study, especially the interaction between forward and backward looking mechanisms.

\citeauthor{RATE} introduce RATE (\citeyear{RATE}) and ELMUR (\citeyear{ELMUR}) as two different approaches to adding long-term memory to DT. RATE stores memory tokens in the context window, read and written mostly through self-attention, whilst ELMUR maintains a separate store of vectors that are read and written through cross-attention. Theoretically, the storage capacity of our approach scales better and is more expressive---we discuss this further in Section \ref{sec:evaluation}.



\section{Methodology} \label{sec:methodology}
We first introduce the Decision Titan (DTi) in Section \ref{sec:methodology_dti}, followed by the X-Maze environment in Section \ref{sec:methodology_xmaze}. For readers not already familiar with TTT \cite{TTT}, Titans \cite{Titans}, and Decision Transformer (DT) \cite{DecisionTransformer}, a full description is provided in Appendix \ref{apx:detailed_preliminaries}.

\subsection{Decision Titan} \label{sec:methodology_dti}

Our main objective is to apply Titans to environments with long-term dependencies in the RL domain. Our key observation is that Titans uses a Transformer backbone, and so can be trained in the same fashion as DT---we use the same trajectory representation, tokenization approach, action head, and training objectives. Where our approach diverges is the use of Titans blocks, the inference approach, the training strategy, and the choice of positional embedding. We use the same chunked inferencing method as Titans.

\subsubsection{Architecture}
Out of the three Titans architectures proposed by \citeauthor{Titans} (\citeyear{Titans}), we utilise \textit{Memory as a Layer} (MAL). At the time of writing, the Titans implementation by \citeauthor{titans-pytorch} (\citeyear{titans-pytorch}) that we build off of only has an implementation for MAL and MAG (Memory as a Gate) architectures. Out of these two, MAL is the simpler architecture and provides a clear separation between long and short-term memory, which could be obfuscated by MAG's sliding attention window. Whilst \citeauthor{Titans} report that MAL has the worst results out of the Titans architectures, it still performs better compared to the other techniques they tested against, making it a suitable architecture to evaluate the potential of TTT in RL.

The Titans MAL block consists of two components, neural memory followed by self-attention. More specifically, for an input
$x_t$, memory module $M$, slow weights $\theta$, and fast weights $\theta'_t$ at time $t$, the order of operations is as follows: 
\begin{align}
    &(\alpha_t, \gamma_t, \eta_t,k_t,v_t,q_t) = f(x_t;\theta)\\
    &s_t=\gamma_t s_{t-1} - \eta_t \nabla \ell(\Vert M(k_t;\theta'_{t-1})-v_t\Vert_2^2)  \\
    &\theta'_{t} = (1-\alpha_t)\theta'_{t-1}+s_t &&\textit{write} \\
    &x_t = M(x_t;\theta'_t) &&\textit{read} \\
    &y = \text{Attn}(x_t). &&\textit{attn}
\end{align}
The self-attention has a size of $3k+n$ tokens, where $k$ refers to the number of timesteps of a sequence the model's short-term memory can see and $n$ refers to the number of persistent memory tokens. In practice, the Titans update occurs only every $b$ tokens and is batched as per \citeauthor{Titans} (\citeyear{Titans}).

\begin{figure}[tp]
    \centering
    \includegraphics[width=1\linewidth]{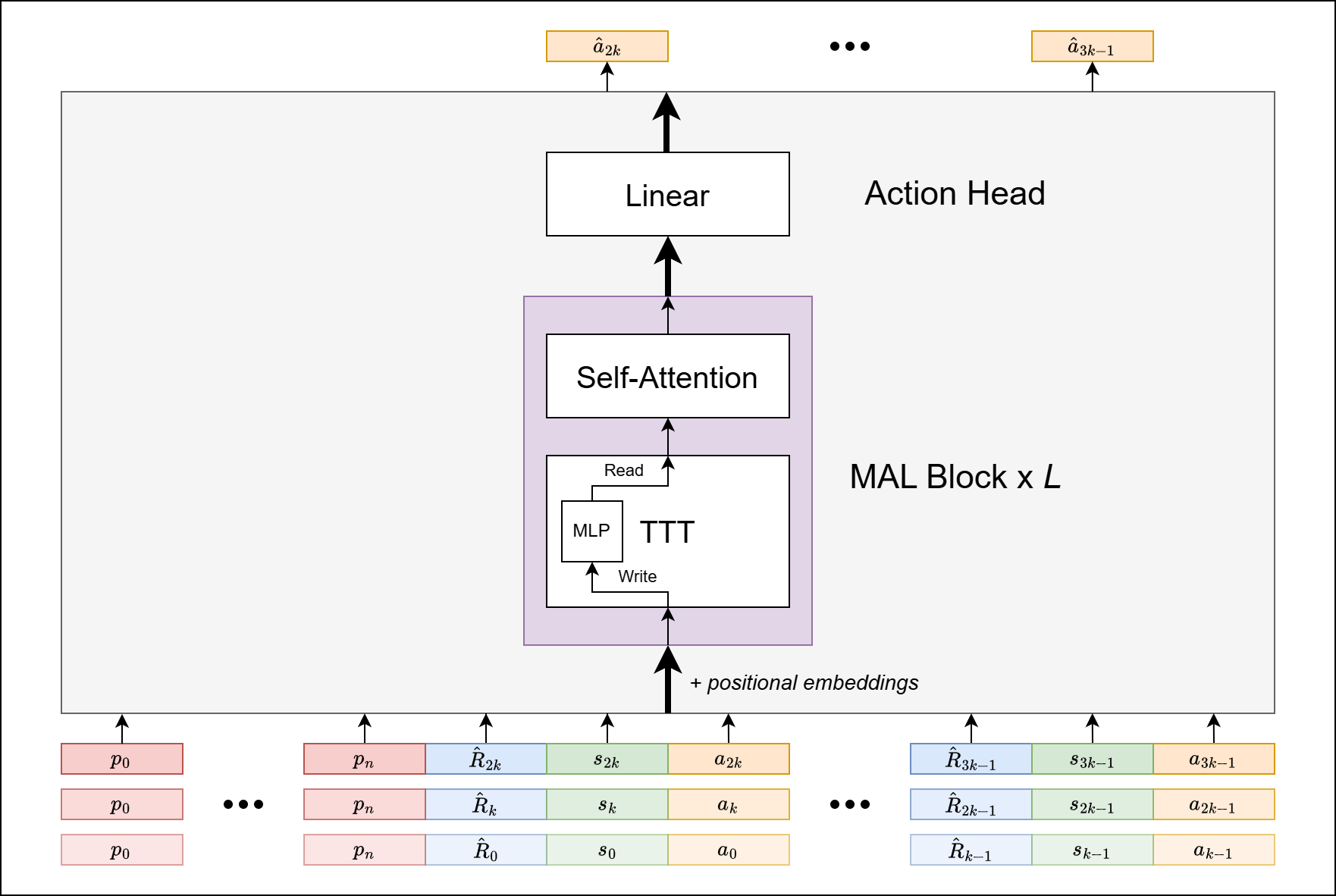}
    \caption{The Decision Titan architecture. A trajectory of returns-to-go, states, and actions is projected into a shared embedding space before having positional embeddings added. This sequence is then chunked into subsequences of $k$ timesteps and $n$ persistent memory tokens are appended to the front of each chunk. Following this, each chunk is then passed through $L$ Titans MAL blocks, consisting of a TTT layer that is updated every $b$ timesteps, followed by self-attention with a window size of $3k+n$ tokens.}
    \label{fig:dti_architecture}
\end{figure}

Whilst $M$ can be any neural network, we utilise a two-layer MLP, as per \citeauthor{Titans} (\citeyear{Titans}) and \citeauthor{TTT} (\citeyear{TTT}). Using different architectures as $M$ marks an interesting direction of research, but it is not the focus of this study---there is further discussion of this in Sections \ref{sec:related_work} and \ref{sec:evaluation}. We choose 2 layers as the experiments of \citeauthor{Titans} suggest that 2 layers makes a significant improvement in results compared to a single layer, whilst further increasing to 3 and 4 only has a small improvement. A full architecture diagram is shown in Figure \ref{fig:dti_architecture}.

\subsubsection{Training}
Due to the long-term memory mechanism of DTi, a different approach to training must be taken compared to DT as randomly sampled length $k$ subsequences do not capture the dynamics of long-range dependencies. To remedy this, we train on batches of entire episode trajectories, padded to the largest length in the batch. Within each trajectory, the sequence is processed in chunks of $k$ timesteps in order. The state of $M$ is not shared between episodes within a batch, nor persisted between training iterations; each trajectory begins with $M$ set to its learnt initial state, $\theta'_0$.

A consequence of training on full trajectories rather than fixed-length subsequences is that memory and compute requirements scale with trajectory length. For environments with very long episodes this may be prohibitive, in which case trajectories could in principle be truncated to a maximum length; however, this is not necessary for the environments considered in this study.

An important point is that there is no backpropagation through time, i.e. there is no flow of gradients between chunks. As a result, calculating the gradients for a whole episode is equivalent to calculating $\lceil T/k \rceil$ individual gradient updates with the correct initial long-term memory state. This leads to an effective batch size of $b\lceil T/k \rceil$.

\subsubsection{Embeddings}
Motivated by how the long-term memory mechanism interacts with different embedding strategies, we test the following 3 strategies. The first is to use sinusoidal embeddings indexed by the timestep $t$. We refer to this as \textit{global embeddings}. The second strategy is \textit{chunked embeddings} which instead index sinusoidal embeddings by the position in a chunk, $t'=t-ck$, meaning each chunk receives the same embedding. The final approach is an \textit{axial positional embedding} \cite{titans-pytorch}. This is a learnt embedding that sums the output of 2 MLPs, one as a function of the intra-chunk position and the other as a function of the chunk, providing DTi with both global and local positional information. A summary of these techniques is provided in Table \ref{tab:embedding_strategies}.


\begin{table}[tp]
    \centering
    \caption{Embedding Strategies}
    \begin{tabular}{c c c}
        \toprule
        Embedding & Learnt & Information \\
        \midrule
         Global & No & Global position\\
         Chunked & No & Local position\\
         Axial Positional & Yes & Global and local position\\
         \bottomrule
    \end{tabular}
    \label{tab:embedding_strategies}
\end{table}

\subsection{X-Maze Environment} \label{sec:methodology_xmaze}

The T-Maze environment \cite{CreditAssignmentVsLongTermDep} is a toy setting used to test the ability of an agent to remember long-term dependencies, featuring a corridor with a T-junction at the end. At the beginning of the corridor the agent will receive a direction, left or right, which the agent then has to turn in at the end of the corridor in order to receive a reward. The key challenge is remembering the information over the length of the corridor. This environment is effective due to its simplicity whilst still isolating the memory function of a model. However, it is limited in that it cannot test the ability of an agent to remember the order of events, nor the ability to remember more than a single bit.

\begin{figure}[tp]
    \centering
    \includegraphics[width=1\linewidth]{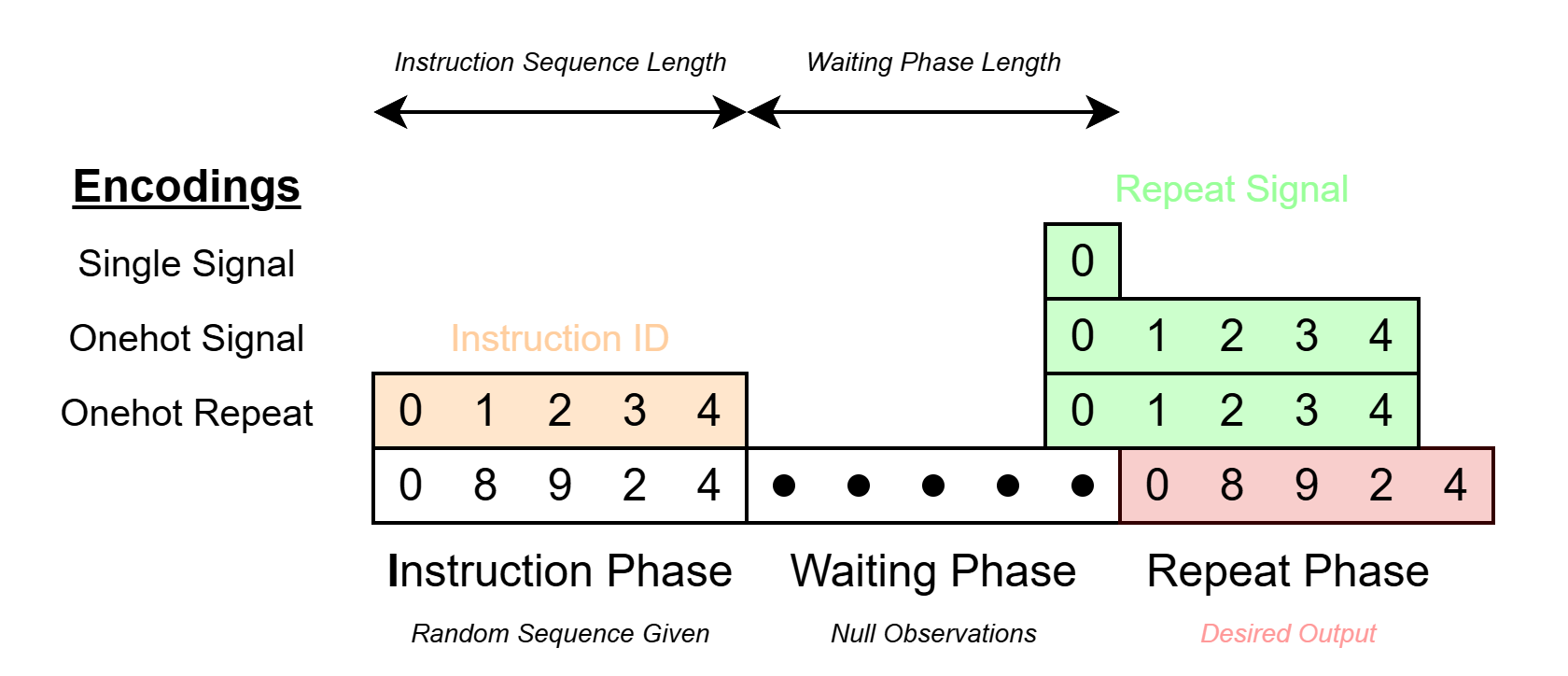}
    \caption{The X-Maze environment. This example shows an instruction phase length of 5, a waiting phase length of 5, and a potential instruction count of 10 (an instruction can be any number from 0 to 9). Time flows from left to right in the diagram. An episode is broken into 3 phases: instruction, waiting, and repeat. There are three different possible environment encodings, which provide different information over the course of an episode.}
    \label{fig:xmaze_diagram}
\end{figure}

We propose the X-Maze environment, an extension of T-Maze. Rather than a single direction, which we refer to from here on as an instruction, an agent must remember and follow a sequence of instructions. Furthermore, the number of possible instructions, $I$, can be set, meaning the environment is no longer restricted to just left or right. As an example, 10 possible instructions and a sequence length of 9 is akin to asking an agent to remember 9 base-10 numbers, e.g. a phone number. A visualisation is shown in Figure \ref{fig:xmaze_diagram}. These changes address the aforementioned issues, testing both memory capacity and sequential memory whilst still remaining simple.

In our implementation, the agent is not moving through a physical space. Instead, the agent has to repeat the given instructions upon receiving a specific signal after a set number of timesteps. The environment was designed this way in order to fully isolate memory capacity, removing any navigation the agent has to perform. This further allows finer control over exactly how many steps the agent has to remember information for, which could vary when an agent can take non-optimal routes. To summarise, an episode can be broken down into 3 phases:
\begin{enumerate}
    \item \textbf{Instruction Phase} ~--~ The agent receives the instructions they later have to repeat.
    \item \textbf{Waiting Phase} ~--~ The agent waits for a period of time until the \textit{repeat signal} is received.
    \item \textbf{Repeat Phase} ~--~ Upon receiving the repeat signal, the agent must repeat the instructions.
\end{enumerate}

The action space of the environment is discrete, with one action corresponding to each possible instruction and one null action representing waiting, making for a total of $I+1$ actions. The reward structure of the environment is dense---the agent receives a -1 penalty every time the agent does not take the null action before the repeat phase or for every instruction it repeats incorrectly, yielding a maximum episode return of 0.

In terms of the observation space, instructions are one-hot encoded in a vector of $I$ dimensions. Furthermore, we employ three strategies of repeat signal that vary in the degree of information provided to the agent about when and what to repeat, each of which modifies the observation space dimensionality accordingly. The first is \textit{single signal}, in which the model only receives a single bit at the last step of the waiting phase. Next is \textit{one-hot signal}, where the agent receives a one-hot encoded id of which instruction to repeat for each step of the repeat phase. Finally \textit{one-hot repeat} which is the same as one-hot signal, except the id is also given during each step of the instruction phase. A summary of these approaches as well as how they modify the observation space is shown in Table \ref{tab:xmaze_encodings}.

\begin{table}[tp]
    \tiny
    \centering
    \caption{X-Maze Encodings}
    \begin{tabularx}{\columnwidth}{c X c}
         \toprule
         Encoding & Description & Obs Dim \\
         \midrule
         Single Signal & A single binary bit is appended to the observation, set to 1 only at the final timestep of the waiting phase to indicate the start of the repeat phase. & $+1$ \\
         \addlinespace
         One-hot Signal & A one-hot vector of dimension $I$ is appended, 
         encoding which instruction the agent should next repeat. This is provided at the last step of the waiting phase and every step of the repeat phase. & $+I$ \\
         \addlinespace
         One-hot Repeat & Identical to One-hot Signal, except the one-hot instruction id is additionally provided during each step of the instruction phase. & $+I$ \\
         \bottomrule
    \end{tabularx}
    \label{tab:xmaze_encodings}
\end{table}

\section{Experiments} \label{sec:experiments}

This section details the experiments conducted to test DTi, as well as the subsequent results and discussion. There are addition experiments provided in Appendix \ref{apx:experiments}, including showing the advantage of DTi compared to DT in T-Maze. We do not test DT in the experiments in this section as it cannot solve them without increasing the context window length. Default hyperparameters for all of our experiments can be found in Appendix \ref{apx:hyperparameters}. When different models are compared, the base set of hyperparameters (e.g. number of layers) are the same unless stated otherwise. When memory is used, a TTT module is used on the 2nd out of 3 layers with a batch size equal to that of the attention window. All environment observations are z-score normalised and we do not use reward scaling. Each X-Maze dataset contains 10,000 trajectories gathered from a perfect hard-coded policy. All tests are ran on a 1/8th partition of a Nvidia A100 GPU.


\subsection{Sequential Memory and Embeddings} \label{sec:experiments_sequential_memory}

We have shown that DTi can remember a piece of information using the long-term memory module, however we have not shown whether it can remember the \emph{order} of events outside the context window. Furthermore, it is unclear how the dynamics of this memory interact with different positional embedding approaches as well as what level of information is required in environment observations for the memory to function. To gain some insight into this question, we test DTi in the X-Maze environment with different combinations of embedding strategies and environment encodings.

\subsubsection{Setup}
We train DTi on a dataset containing instruction sequence lengths in the range of [6,10], waiting period lengths in the range of [6,10], and a possible instruction count of 10. The sequence length and waiting period are varied to prevent the model from figuring out when the repeat phase is through global token position. As the instruction sequence length is greater than 1, maximal performance in this setting requires remembering the order in which instructions were received. At testing, the model is evaluated on an X-Maze setup matching the most challenging case of the training data---an instruction sequence length and waiting period length of 10. All 9 pairwise combinations between the 3 X-Maze environment encodings outlined in Section \ref{sec:methodology_xmaze} and the 3 embedding strategies outlined in \ref{sec:methodology_dti} are tested.

\subsubsection{Results and Discussion}

The mean return of each of the 9 combinations are displayed in Table \ref{tab:seq_mem_results}. The results show that the model's performance depends on the environment encoding, but not on embeddings as the performance of the model is the same across embeddings for each environment encoding. DTi achieves perfect results in the one-hot repeat setting, and close to perfect in one-hot signal. However, a steep drop in mean return is observed in the single signal setting, achieving mean scores only in the range of $[-5.76,-5.6]$, although this is still better than random chance (-9).

To understand these results, we must recall how the long-term memory module works. Given timesteps $t, t'$ where $t' < t$, the query at time $t$ must approximate the key from time $t'$ in order for information to be recalled from time $t'$. In essence, the model must learn to associate queries to keys of earlier timesteps. Remember that these queries and keys are projections of the input, $(q_t,k_t)=W(x_t)$, and it is the projection matrix that is learnt over training. 

In the one-hot setting, there is an easy association to make between the instruction id given alongside each instruction and the same instruction id given as the remembering signal, explaining the perfect score across embeddings. In the one-hot signal setting, the instruction id given as the remembering signal can be associated with the positional embedding of the instructions since the instructions always appear in the first 10 timesteps---instruction 0 always appears in position 0. In the single signal setting, there is no clear signal during the remembering phase that can be associated with the positional embedding of the instructions, leading to the decline in performance.

Global positional information of tokens does not help due to the randomness in the waiting period length. For example, sometimes the model may be asked to recall the first instruction at $t=21$, other times at $t = 24$, and so there is no single instruction that a timestep can be associated with. This explains why all embedding strategies performed similarly.

This decisively answers the question we set out. The memory module \textit{can} remember sequential information on the condition that the environment provides cues that can be reliably associated with the information that must be recalled. However, the long-term memory module struggles to remember the order of events outside of this, highlighting a limitation of the technique and a potential direction for future work.

\begin{table}[tp]
    \tiny
    \centering
    \caption{Mean Return of Embeddings/Encoding Combinations}
    \begin{tabular}{c | c c c}
        \toprule
        Encoding & Chunked Embedding & Global Embedding & Axial Embedding  \\
        \midrule
        Single Signal & $-5.6\pm1.81$ & $-5.72\pm 2.11$  & $-5.76\pm1.53$ \\
        One-hot Signal & $-0.16 \pm 0.46$ & $-0.20 \pm 0.49$ & $-0.72\pm1.28$ \\
        One-hot Repeat & $-0.00 \pm 0.00$ & $-0.00 \pm 0.00$ & $-0.00 \pm 0.00$ \\
        \bottomrule
    \end{tabular}
    
    \label{tab:seq_mem_results}
\end{table}

\subsection{Heads} \label{sec:experiments_heads}

In this test, we examine the effect of increasing the number of MLP heads in the architecture of the neural memory model $M$. We refer to heads in this context as the number of different copies of the model running in parallel, i.e. a head count of 2 for $M$ would mean having 2 two-layered MLPs in parallel, doubling the number of parameters in the neural memory module compared to a single head.

\subsubsection{Setup}
 We test DTi in 3 different setups of the X-Maze environment with waiting phase lengths of [11,15], [46,50], and [96,100]. Each length is trained for 2500, 5000, and 15000 batches respectively. All 3 setups use the one-hot repeat encoding, an instruction sequence length range of [6,10], and a possible instruction count of 10. For each environment setup, we test DTi with each of the 3 embedding strategies outlined in Section \ref{sec:methodology_dti} and with 1, 2, and 4 MLP heads in the neural memory module. This amounts to 27 total configurations, and we do 5 runs on different seeds in each configuration. Each run is evaluated every 250 batches on 5 different seed in the hardest environment of each configuration. 

\subsubsection{Results and Discussion}

We report the fraction into training it takes before DTi achieves a mean episode return of 0 in Table \ref{tab:heads_result}, averaged across runs and embeddings. For example, a score of 0.4 when the model is being trained for 2500 batches would indicate that DTi needs 1000 batches of training on average until it achieves a perfect score in the environment. In the event that a run never achieves a score of 0, it is counted as taking the whole training period. Each cell in table is the average of 15 different runs---5 seeds times 3 embeddings.

\begin{figure}[tp]
    \centering
    \includegraphics[width=1\linewidth]{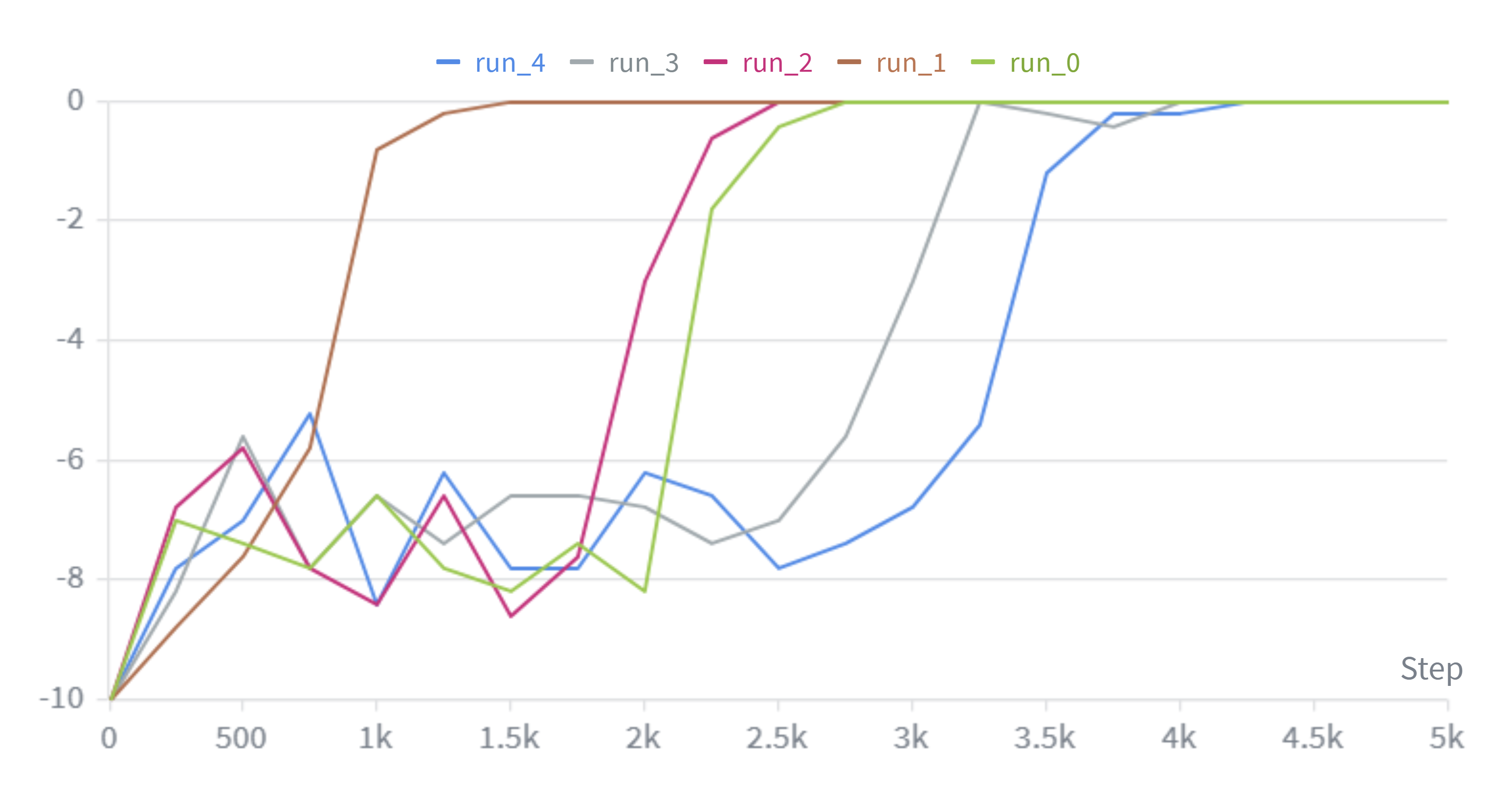}
    \caption{Mean episode return of 5 DTi runs against training steps. Each run is evaluated every 250 batches by sampling 5 episodes in the same environment used to gather training data. In this case, the environment is X-Maze with a [11,15] waiting phase range with DTi using the axial embedding strategy and a single head in $M$. All runs rapidly achieve a mean return equivalent to random chance, before at varying points rapidly increasing to 0 once the long-term dependencies are learnt.}
    \label{fig:heads_training_example}
\end{figure}

The results show that increasing the number of heads consistently decreases the amount of training it takes for the dependencies to be learnt. Furthermore, we observe that there is no in-between---either the model learns the long-term dependencies or it does not. An example is shown in Figure \ref{fig:heads_training_example} in which 5 runs fluctuate around the mean return of a random policy, before each run rapidly increases to an episode return of 0 at different points. This same pattern is seen across all 27 configurations.

These results can be explained by the flow of gradients in the Titans architecture. As there is no back-propagation through time, gradients do not pass between chunks and there is no flow between the end points of a dependency. This suggests that Titans memory, and indeed the RNN proposed by \citeauthor{TTT} \cite{TTT}, learns dependencies through chance patterns in $M$ caused by the required observations earlier in the sequence. Increasing the number of heads increases the number of chance patterns available to model, and in turn the odds that the model is provided a pattern it can pickup and learn from. We provide further evidence for our hypothesis in Section \ref{sec:experiments_gate_visualisation}. This explains the observed results and provides a deeper depth of understanding to this style of TTT memory.

\begin{table}[tp]
    \centering
    \caption{Mean First 0 Episode Return}
    \begin{tabular}{c | c c c}
        \toprule
        Heads & [11,15] & [46,50] & [96,100] \\
        \midrule
        1 & 0.57 & 0.61 & 1.00 \\
        2 & 0.42 & 0.38 & 0.81 \\
        4 & 0.35 & 0.37 & 0.71 \\
        \bottomrule
    \end{tabular}
    \label{tab:heads_result}
\end{table}

%

\subsection{Out-Of-Training Generalisation} \label{sec:experiments_generalisation}

In this test, we examine how well DTi generalises to environment lengths outside of the training data, as well as how different positional embedding strategies effect this.

\subsubsection{Setup}

We train DTi on 5 seeds in 3 X-Maze environments with different ranges of waiting phase lengths; [11,15], [46,50], [96,100]; which also determines the number of batches the model is trained for: 2500, 5000, 15000. Otherwise, the 3 environments are the same, having the same instruction sequence length of 10, possible instruction count of 10, and one-hot repeat environment encoding. For each of the 3 environments, DTi is tested with the 3 embedding strategies outlined in Section \ref{sec:methodology_dti}, making for a total of 9 configurations. Each configuration is then evaluated on instruction lengths of 10 (longest seen in training data) on waiting phases in the range of [5,200], incremented in steps of 5. 4 heads are used on the memory module in order to improve the success rate of the model learning dependencies. 

\begin{table}[tp]
    \centering
    \caption{Number of Successful Runs}
    \begin{tabular}{c|c|c|c}
    \toprule
         Embedding & [11,15] & [46,50] & [96,100] \\
         \midrule
         Chunked Embedding & 5 & 5 &  2 \\
         Global Embedding & 5 & 5 & 1 \\
         Axial Embedding & 5 & 5 & 5 \\
         \bottomrule
    \end{tabular}
    \label{tab:falloff_success_count}
\end{table}

\subsubsection{Results and Discussion}

\begin{figure}[tp]
    \centering
    \includegraphics[width=1\linewidth]{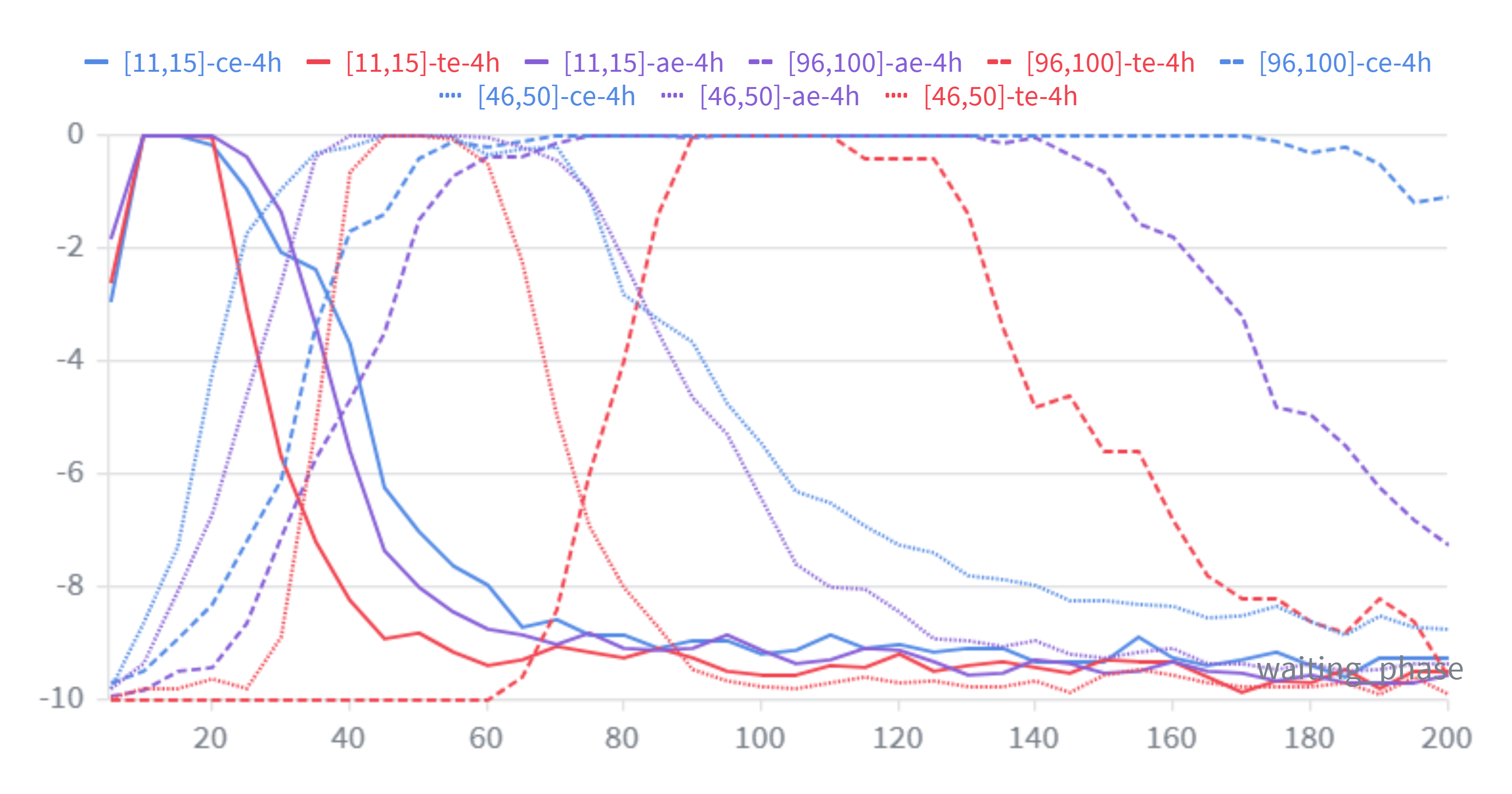}
    \caption{Average episode return of the repeat phase (last 10 environment steps) against waiting phase length, comparing different DTi versions in the X-Maze environment. Trained was performed on 3 different waiting phase length ranges of X-Maze: \textbf{[11,15]}, \textbf{[46,50]}, and \textbf{[96,100]}. 3 different embeddings are tested for each length, \textbf{chunked embeddings (ce)}, \textbf{global embeddings (te)}, and \textbf{axial positional embeddings (ae)}. The results show that DTi can generalise to lengths of at least 1.7x what it was trained on and that chunked embeddings consistently generalise to greater lengths than the other embedding strategies. Interestingly, a model succeeding at a given length does not mean it can succeed at a shorter length. Each data point is the average of at least 5 episodes.}
    \label{fig:memory_falloff}
\end{figure}

The average mean return of the repeat phase for each of the 9 configurations across different waiting phase lengths is shown in Figure \ref{fig:memory_falloff}. Each point on the graph is the average of 5 episodes per 5 training seeds, making for an average over 25 episodes, except for the seeds in which the model failed to learn the dependencies. In such an instance, the run is ignored---a summary of the number of successful runs per configuration is shown in Table \ref{tab:falloff_success_count}.

The results show that DTi can generalise to waiting phases of at least 1.7x what it was trained on---DTi with a chunked embedding trained on the [96,100] range achieves perfect mean returns up until lengths of 175. The next point to note is that chunked embeddings generalise the most, followed by axial embeddings, with global embeddings generalising the least. Finally, the curve of each configuration has a rise at the beginning, not just a falloff at the end, showing that being able to memorise a value for a given length does not necessarily mean it is available straight away.

These observations can once again be explained by examining the mechanisms of the memory module. Recalling the memory update and access rules from Equations \ref{eq:titans_update} and \ref{eq:ttt_access_rule}, the model queries the memory by inferencing on a learnt projection of the input. When using global embeddings, this projection matrix receives embeddings at longer episode lengths that it has not been trained on and does not understand. This explains why chunked embeddings generalise the best, as the model never sees embeddings that it has not already seen at training. 

Despite this, the memory still eventually fails on the chunked embedding. Furthermore, the additional number of timesteps the model can generalise too increases relative to the lengths seen in the training data. A potential explanation is the forgetting gate in the update rule. By default, the model will not minimise the forgetting gate as there is no requirement to do so, causing memory to decay. When the training data contains only shorter-range dependencies, the data does not need to be remembered for long, putting less pressure on the model to minimise the value of the forgetting gate compared to when the dependencies are of longer range. Therefore, the memories of the models that were trained on shorter ranges of waiting phases in X-Maze degrade quicker due to larger forgetting gates. We show that the most likely overall limiting factor to generalisation is catastrophic forgetting in Section \ref{sec:experiments_gate_visualisation}, although this does not explain the relative increase of the generalisation length.

As to why the memory is not guaranteed to generalise to shorter lengths, the weights of the memory module are constantly shifting from the momentum of prior observations and the losses of new ones. Given an observation that must be recalled later, the loss that it produces in the memory module is applied repeatedly over timesteps, building strength over time. If not enough time has passed, then the strength of this signal may not be strong enough to be recognised. 



\subsection{Gate Visualisation and Momentum Ablation} \label{sec:experiments_gate_visualisation}

We have shown that the Decision Titan can learn and remember long-term dependencies, however the question remains as to how the memory module learns in practise.

\subsubsection{Setup}

We run DTi in the X-Maze environment and visualise at every timestep $t$: the forget gate $\alpha_t$, the average magnitude of the momentum $s_t$, the momentum gate $\gamma_t$, the write gate $\eta_t$, and the loss $\mathcal{L}$. See Equation \ref{eq:titans_update} for reference. Furthermore, we compare these values to DTi without momentum, leading to an update rule closer to the formulation of \citeauthor{TTT} \cite{TTT} in Equation \ref{eq:ttt_update_rule}, containing only $\alpha_t$, $\eta_t$, and $\mathcal{L}$. For both the momentum and no momentum scenario, 5 runs were trained on waiting phase length ranges of [46,50] and [96,100]---a sequence length of 9 and possible instruction count of 10. In all 4 of these configurations, we use DTi with chunked embeddings for the best generalisation and 4 heads to verify the idea that more heads provide more chances for the model to learn long-term dependencies.

\begin{figure}[tp]
    \centering
    \includegraphics[width=1\linewidth]{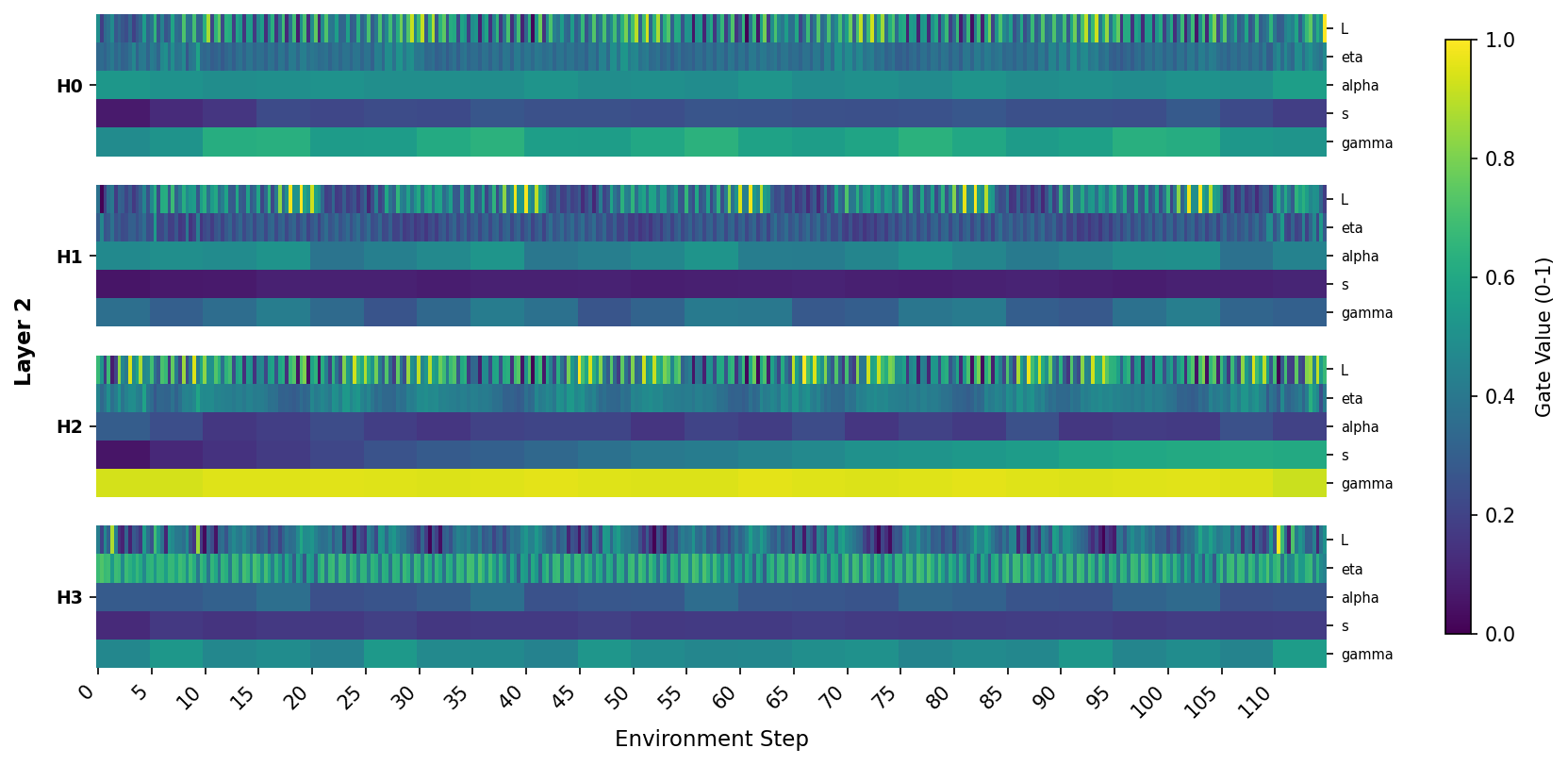}
    \caption{Visualisation of a successful run in X-Maze with a waiting phase of length 100. L refers to the loss $\mathcal{L}$ at time $t$, eta to the write strength $\eta_t$, alpha to the forgetting gate $\alpha_t$, s to the momentum $s_t$, and gamma to the momentum decay $\gamma_t$. L is [0,1] normalised, H$X$ refers to head number $X$.}
    \label{fig:gates_100_success}
\end{figure}

\begin{figure}[tp]
    \centering
    \includegraphics[width=1\linewidth]{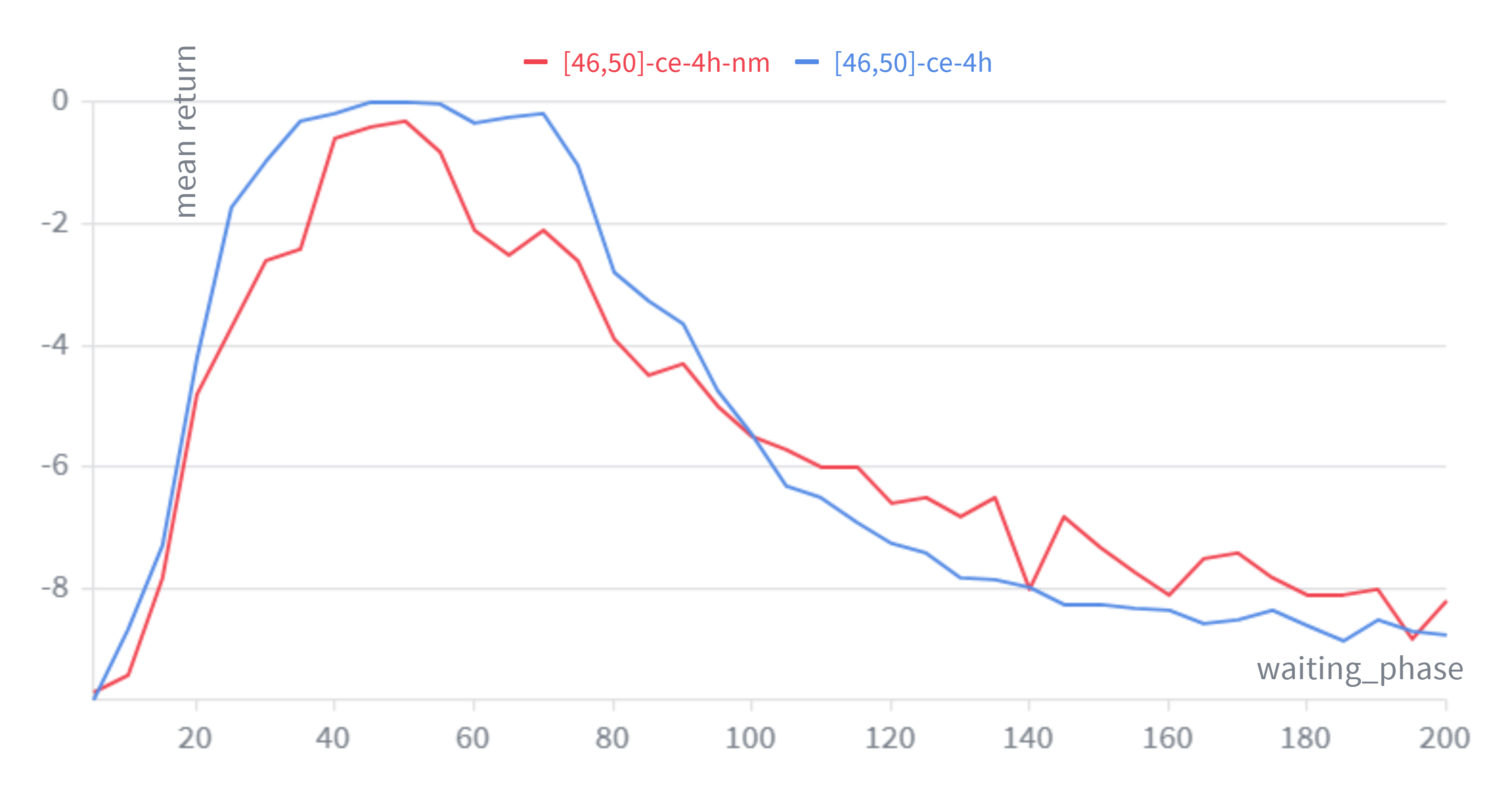}
    \caption{Mean return across different waiting phase lengths. The red run is without momentum \textbf{(nm)} and blue is with. 2 runs are averaged in the no momentum scenario, whilst with momentum is the average of 5 runs, and each run is evaluated on 5 different seeds for each point. One of the runs without momentum was suboptimal due to under-training, explaining the slightly lower peak of the no-momentum case.}
    \label{fig:momentum_ablation_falloff}
\end{figure}

\subsubsection{Results and Discussion}

An example of a successful run in the environment with a 100 length waiting phase is shown in Figure \ref{fig:gates_100_success}. We describe all relevant observations, however additional examples are provided in Appendix \ref{apx:experiments_gate_visualisation}.

In all cases using momentum where the model successfully learns the task, exactly one head learns to maintain a high value of $\gamma$ across the episode---around $0.95$. This means previous gradients decay slowly and the momentum, $s_t$, builds over time. This does not occur in failure cases, suggesting that momentum has an important part to play in the model's success. This is confirmed by the fact that without momentum, the model's success rate in learning the dependencies was 0\% in waiting phases of both 50 and 100, compared to 100\% and 40\% with momentum.

The model also learns a low value for the forgetting gate, but not as low as expected. For example, the run shown in Figure \ref{fig:gates_100_success} maintains $\alpha \approx 0.3$. This would mean that the effect on the parameters of $M$ from storing a value in the first chunk would only retain $(1-\alpha )^{21}\approx 0.056\%$ of its magnitude by the time it needs to be recalled. It is therefore very unlikely that the main model would be able to reconstruct the stored value from this weak signal when $M$ is queried, showing that momentum helps \emph{propagate} dependencies.

This is verified by comparing successful cases with and without momentum. We extend the training of the model without momentum to 6000 in the waiting phase length 50 setting in order to get 2 successful cases, both of which end up having a much lower $\alpha$ of around 0.05. This shows that when information cannot be propagated through momentum, the model is forced to learn to discard less information at each timestep. This reveals how momentum actually improves results---it helps persist long-term dependencies, making it easier for the model to learn them. 

This raises another question---does momentum effect the ability of a model to generalise? Given that using no momentum forces the model to learn a smaller forget gate, this will potentially help preserve information through longer, unseen environment lengths. We take the 2 successful no momentum runs as well as the 5 regular runs and plot how they decay, following the same procedure as Section \ref{sec:experiments_generalisation}. The results in Figure \ref{fig:momentum_ablation_falloff} show that the decay in performance is unchanged, implying that the forgetting gate is not the limiting factor for longer generalisation. However, the weights of the memory module are slightly changed at every chunk through the update rule, even when there is no relevant information to store, potentially causing degradation of memory over time. This leaves catastrophic forgetting as the most likely remaining cause limiting the ability of the model to generalise, considering the embeddings can be ruled out as the chunked embeddings are the same for every chunk. We discuss the implications of this fully in Section \ref{sec:evaluation}.

Finally, the above observations and the fact that the model only learns to propagate information through a single head confirms the hypothesis from Section \ref{sec:experiments_heads} that heads provide more chances for long-term dependencies to be propagated far enough to be learnt. If it was a question of memory capacity, more than one head would have learnt to maintain momentum or to minimise the forgetting gate. 

\section{Discussion} \label{sec:evaluation}

Overall, the DTi architecture displays a clear advantage over a standard Decision Transformer. Whilst DT's scaling is quadratic, we have shown that the long-term memory of DTi can learn long-term dependencies with a range of at least 20x the context window length and can extrapolate to 1.7x the episode lengths seen at training. DTi also improves upon the scaling of memory capacity compared to previous approaches in Offline RL. The time complexity of the memory read and write operations is constant with respect to sequence length and scales linearly with respect to the number of parameters in the memory module. This makes for a total model complexity of $O(k^2+p)$, where $k$ is the context length and $p$ is the number of parameters in the MLP memory module. The time complexity of RATE \cite{RATE} is $O((k+m)^2)$, where $m$ is the number of memory tokens, and ELMUR's \cite{ELMUR} complexity is $O(k^2+km)$---these models were previously discussed in Section \ref{sec:related_work}. Furthermore, as parameters compress information more than tokens, $p$ MLP parameters can store more information than $m$ tokens. However, this is only in theory and we do not directly compare DTi to RATE and ELMUR, or how their storage scales in practice, marking a limitation of our study and an interesting direction for future work.

With that said, DTi has some limitations. The strategy of training on entire episodes constrains the batch size, is unsustainable for longer environments, and slows down training compared to DT. The dependency of a chunk on the memory state from the previous chunk, as well as the initial memory state being a learnt slow weight, complicate the design of an efficient training setup. We have also demonstrated that DTi has limited ability to remember sequential information, but this could potentially be solved through the TTT framework with a different memory module or through relative positional embeddings \cite{RelativeEmbeddings,RoPE}. These are both limitations of our methodology, and could be the focus of future research. 

A notable limitation is the gap between the ability of DTi and ELMUR to extrapolate. Whilst we do not directly compare the models, ELMUR has been shown to generalise to T-Maze corridor lengths of at least 1,000,000 steps \cite{ELMUR}, whilst DTi falls off at 200 steps in X-Maze. However, we believe the Test-Time Training framework could close this gap with the right memory module, based on the following. We have shown that the most likely limiting factor in DTi's ability to extrapolate is catastrophic forgetting, caused by corruption over time from gradient updates every chunk. The generalisation of RATE, which updates all memory tokens each chunk, is also poor compared to ELMUR's. ELMUR achieves better results through sparse memory updates---only the single least recently used token is updated at the end of a chunk, meaning all other tokens are untouched and uncorrupted. Such sparse access schemes are also possible under the TTT framework---for example FwPKM \cite{FastWeightProductKeyMemory}, the time complexity of which is only $O(k^2+\sqrt{m})$. Therefore, we believe applying such a scheme to RL to be the most promising direction for further study.

We mechanistically explain three phenomena: the role of momentum in Titans memory updates, the necessity of propagating long-term dependencies during training, and the influence of memory heads on their propagation. This provides a foundation future work can build on to extend the length of this propagation, for example, through scheduled biasing of the update rule gates during training. Other potential directions include combining DTi with other, orthogonal DT extensions, looking at the effect of persistent memory tokens, and testing DTi in noisier and more demanding environments such as MemoryMaze \cite{MemoryMaze} or the MIKASA-Robo suite \cite{MIKASA-Robo}. 

\section{Conclusion} \label{sec:conclusion}

In this paper, we have introduced Decision Titan by merging Decision Transformer with Titans, capable of tackling RL environments with long-term dependencies. We also introduce the X-Maze environment, a controlled environment capable of testing the capacity of a model to remember the order of long-term dependencies over large time horizons, a capability that is crucial for sequential decision making.

Overall, this work demonstrates that the Test-Time Training framework is a viable foundation for long-term memory in Offline RL, and that there remains significant room to improve upon it. We hope that the mechanistic insights provided here, as well as the X-Maze environment, will serve as useful tools for future research in this direction.

\section*{Impact Statement}
This paper presents work whose goal is to advance the field of Machine Learning. There are many potential societal consequences of our work, none which we feel must be specifically highlighted here.



\bibliography{bibliography}
\bibliographystyle{icml2026}

\newpage
\appendix

\section{Detailed Preliminaries} \label{apx:detailed_preliminaries}

We first outline the TTT framework, that allows an arbitrary neural network to function as a long-term memory module, followed by Titans, an architecture which combines TTT with Transformers. We follow by detailing the Decision Transformer (DT) architecture that applies Transformers as a policy learner in the Offline RL domain. 

\subsubsection{Test-Time Training} \label{sec:methodology_background_ttt}

An RNN maintains a hidden state, $h_t$, and has an update and output rule. The dynamics of an RNN with input $x_t$, output $y_t$, and parameters $\theta$ can be represented as follows: 
\begin{gather}
    h_{t}=f(x_t,h_{t-1};\theta) \\
    y_t=g(x_t,h_t;\theta)
\end{gather}
In a standard RNN, $h_t$ is a vector whose value is updated over time, e.g. $h_t=\sigma(W_1h_{t-1}+W_2x_t)$. However, these hidden states do not have to be a vector. For example, self-attention can be interpreted as an RNN whose hidden state is a linearly growing list of key-value pairs \cite{TransformersAreRNNs, TTT}, where $v_t,k_t,q_t$ are linear projections of $x_t$:

\begin{gather}
    h_{t}=\{h_{t-1},(k_t,v_t)\} \\
    y_t= \text{softmax}(\frac{q_tK_t^T}{\sqrt{d_k}})V_t
\end{gather}

This leads to a more expressive hidden state but the computational cost scales quadratically with respect to the input length. The idea behind the TTT framework is to make the hidden state a set of parameters, referred to as fast weights, of an arbitrary neural network---the memory module $M$. By training $M$ in an 'inner-loop' on a self-supervised reconstruction objective at both training and inference, the model learns to compress a sequence into $h_t$:

\begin{gather}
    h_t = h_{t-1}-\nabla \ell(\Vert M(x_t;h_{t-1})-x_t\Vert_2^2) \\
    y_t = M(x_t;h_t)
\end{gather}

However, this objective is insufficient for a long-term memory as the output of the module is trained to be identical to the input, meaning no new information can be retrieved. Furthermore, there is no way for the model to choose what to remember or forget, leading to the storage of redundant information. These issues can be remedied by introducing linear projections similar to self-attention and an input-dependent gate on the loss, all learnt as part of $\theta$ in the outer loop. This yields the formulation used by \citeauthor{TTT} (\citeyear{TTT}):

\begin{gather}
    (\eta_t, k_t, v_t, q_t) = f(x_t;\theta) \\
    h_t = h_{t-1}-\eta_t\nabla \ell(\Vert M(k_t;h_{t-1})-v_t\Vert_2^2) \label{eq:ttt_update_rule} \\
    y_t = M(q_t;h_t) \label{eq:ttt_access_rule}
\end{gather}

$\eta_t$ modulates the magnitude of the gradient update, effectively acting as a learnt write gate that controls how strongly, if at all, a given input is stored in memory. Crucially, $\eta_t$, $h_0$, and the aforementioned projections are learnt in the outer loop, acting as hyperparameters to the inner loop. The result is a long-term associative memory module with $O(1)$ storage and retrieval complexity with respect to the sequence length, assuming $M$ is stateless. However, its success relies on $M$ learning and maintaining the relationship between $k_t$ and $v_t$, and the outer loop learning the projections such that it links queries to the keys of relevant prior information. If successful, given a time $\tau < t$ and $\theta$ such that $q_t \approx k_\tau$, then $v_\tau$ can be retrieved at time $t$.

This approach is effective, yet exhibits the characteristic RNN limitation of limited parallelization due to the sequential dependency of the update rule. To mitigate this, \citeauthor{TTT} (\citeyear{TTT}) propose breaking the input into batches of size $b$, calculating every inner loop gradient update in a batch using the same stale set of parameters from the end of the last batch:

\begin{gather}
    t' = t - (t \text{ mod } b) \\
    h_t=h_{t'}-\eta_t \sum_{\tau=t'}^{t} \nabla\ell (h_{t'};x_\tau) \label{eq:ttt_batched_update}
\end{gather}

Referred to as \textit{mini-batch gradient descent}, this creates a trade-off where increasing $b$ reduces runtime at the cost of lower fidelity gradient steps caused by calculating updates using stale states. 

\subsubsection{Titans} \label{sec:methodology_background_titans}

Motivated by the observation that the RNN with TTT layers introduced by \citeauthor{TTT} (\citeyear{TTT}) only has a long-term memory, \citeauthor{Titans} (\citeyear{Titans}) add TTT layers to a Transformer backbone to get the advantages of both a short and long-term memory. The input sequence is broken up into chunks, with inter-chunk information being propagated through the memory module. They propose 3 different ways to add the memory module, which can be summarised as follows:

\begin{itemize}
    \item[] \textbf{Memory as a Context} (MAC)---Output of the memory module is appended to a residual connection before being passed through self-attention, providing additional context.
    \item[] \textbf{Memory as a Gate} (MAG)---The memory module is run in parallel to sliding self-attention, the output of which is then used to gate the attention output through an element-wise product.
    \item[] \textbf{Memory as a Layer} (MAL)---The memory module is simply another layer, followed by self-attention.
\end{itemize}

Noticeably, whilst self-attention is used, these are not Transformer blocks as no feed-forward layers are used after the attention. Instead, Titans utilises persistent memory tokens - a learnt set of tokens that are appended to the beginning of each chunk. The idea is that these tokens encode persistent information about the task and \citeauthor{Titans} demonstrate how this produces a bias similar to what is induced when a feed-forward network is used with softmax instead of RELU as an activation \cite{Titans}.

Another change is that Titans introduces a forgetting gate and momentum to the update rule of the memory module. These components are once again input-dependent and learnt in the outer loop of the model, allowing the model to better manage memory:

\begin{gather}
    (\alpha_t,\gamma_t,\eta_t, k_t, v_t, q_t) = f(x_t;\theta) \\
    h_t = (1-\alpha_t)h_{t-1}+s_t \label{eq:titans_update} \\
    s_t = \gamma_t s_{t-1}-\eta_t\nabla \ell (\Vert M(k_t;h_{t-1})- v_t \Vert_2^2)
\end{gather}

This update rule can be chunked in the same fashion as Equation \ref{eq:ttt_batched_update}. Through ablation studies and testing on the Needle-In-A-Haystack benchmarks \cite{RULER}, they show that these changes improve performance compared to TTT in NLP tasks with long term dependencies. 

\subsubsection{Decision Transformer} \label{sec:methodology_background_dt}
DT \cite{DecisionTransformer} seeks to utilise the Transformer architecture to learn a policy in the offline setting, in which a policy is learnt from an offline dataset of trajectories generated from some other policy. As trajectories are simply sequences of states, actions, and rewards, it is possible to cast policy learning as a sequence modelling problem---a setting in which Transformers excel. 

\citeauthor{DecisionTransformer} (\citeyear{DecisionTransformer}) achieve this by training a Transformer to predict the next action in a sequence, conditioned on the \textit{returns-to-go}. Defined as the cumulative future reward from timestep $t$,
\begin{gather}
    \hat{R}_t = \sum_{t' = t}^{T} r_{t'},
\end{gather}
conditioning the model on the returns-to-go allows a desired episode return to be specified. This conditioning is made possible by representing trajectories as an interleaved sequence of returns-to-go, observations, and actions,
\begin{gather}
    \tau = (\hat{R}_0, o_0, a_0, \ \hat{R}_1, o_1, a_1, \ \ldots, \ \hat{R}_{T-1}, o_{T-1}, a_{T-1}), \label{eq:dt_trajectory}
\end{gather}
effectively turning policy learning into a conditional sequence generation problem.

DT adopts a causal Transformer architecture similar to GPT, where tokens are processed autoregressively using masked self-attention. The model uses a context window of $k$ timesteps, meaning the model attends to the last $3k$ tokens in the sequence at each timestep, enabling it to capture long-range dependencies within its context. Each modality (returns, observations, actions) is embedded into a shared latent space through linear projectors for vector inputs and CNNs for visual inputs. In the original formulation, \citeauthor{DecisionTransformer} (\citeyear{DecisionTransformer}) then augment this embedding with positional encodings based on the \textit{episode timestep} before being passed through the Transformer layers. This means that the 3 tokens at each timestep share the same positional embedding.

To predict actions, the hidden representation is passed through a linear projection referred to as the action head:
\begin{gather}
    \hat{a}_t = f_{\theta}(\hat{R}_{t-k+1}, o_{t-k+1}, a_{t-k+1}, \dots, \hat{R}_t, o_t).
\end{gather}
In practice, DT is trained with a classification objective for discrete action spaces and a regression objective with a mean-squared error loss for continuous ones. For efficiency, the model is trained using fixed-length subsequences of $k$ timesteps sampled from trajectories, rather than whole trajectories.

DT demonstrates that sequence modelling can be used to learn competitive policies in a range of offline RL benchmarks, including Atari and Mujoco. Notably, it avoids many of the instabilities associated with value function estimation and bootstrapping, instead relying purely on supervised learning over trajectories. However, it relies on large high-quality datasets and is limited by the size of the context window, the latter of which we aim to address by combining DT with Titans.

\section{Hyperparameters} \label{apx:hyperparameters}

This section includes all of the hyperparameters for our experiments in Table \ref{tab:hyperparameters}.

\begin{table}[h]
    \centering
    \caption{Hyperparameters}
    \label{tab:hyperparameters}
    \begin{threeparttable}
        \begin{tabular}{lll}
            \toprule
            \textbf{Hyperparameter} & \textbf{MuJoCo} & \textbf{X-Maze}\\
            \midrule
            \multicolumn{2}{l}{\textbf{Training}} \\
            \midrule
            \quad \texttt{num\_batches}               & 40001 & - \\
            \quad \texttt{batch\_size}     & $2^{**}$ & 16 \\
            \quad \texttt{learning\_rate}             & $1 \times 10^{-4}$  & $1 \times 10^{-4}$ \\
            \midrule
            \multicolumn{2}{l}{\textbf{Model Architecture}} \\
            \midrule
            \quad \texttt{num\_layers}        & 3 & 3 \\
            \quad \texttt{dimensions}         & 128 & 32 \\
            \quad \texttt{context\_len}       & 20 & 5 \\
            \quad \texttt{heads}              & 1 & 1 \\
            \quad \texttt{num\_persist\_tokens}  & 9 & 6 \\
            \midrule
            \multicolumn{2}{l}{\textbf{Model Settings}} \\
            \midrule
            \quad \texttt{is\_discrete}     & False & True \\
            \quad \texttt{embedding\_type}  & chunked & - \\
            \midrule
            \multicolumn{2}{l}{\textbf{Neural Memory Architecture}} \\
            \midrule
            \quad \texttt{dimensions} & 128 & 32 \\
            \quad \texttt{heads}      & 1 & 1 \\
            \quad \texttt{layers}     & 2 & 2 \\
            \quad \texttt{expansion\_factor} & 4 & 4 \\
            \midrule
            \multicolumn{2}{l}{\textbf{Neural Memory Settings}} \\
            \midrule
            \quad \texttt{max\_lr}      & 1 & 1 \\
            \quad \texttt{batch\_size}  & 69 & 21 \\
            \midrule
            \multicolumn{2}{l}{\textbf{Logging \& Evaluation}} \\
            \midrule
            \quad \texttt{train\_seed}       & [42-46] & [42-46] \\
            \quad \texttt{eval\_seed}        & [0-4] & [0-4] \\
            \quad \texttt{checkpoint\_every} & 2500 & 250 \\
            \quad \texttt{generate\_every}   & 2500 & 250 \\
            \quad \texttt{eval\_episodes}    & 5 & 5 \\
            \quad \texttt{target\_returns}   & * & 0\\
            \bottomrule
        \end{tabular}
        \begin{tablenotes}
            \footnotesize
            \item * Targets of 6000, 3600, and 5000 are used for HalfCheetah, Hopper, and Walkerd2D respectively. ** Effective batch size of 100.
        \end{tablenotes}
    \end{threeparttable}
\end{table}

\section{Additional Experiments} \label{apx:experiments}

\subsection{T-Maze} \label{sec:experiments_t-maze}

In this experiment, DTi is tested against standard DT in a minimal T-Maze. The purpose of this test is to demonstrate that TTT memory works in an RL context, allowing DTi to solve tasks that DT cannot with the same size context window.

\subsubsection{Setup}

To this end, we test DTi and DT in the single signal X-Maze environment with an instruction sequence length of 1 and a possible instruction count of 2, resulting in a setup equivalent to a T-Maze. We set the waiting phase to 5 timesteps and the context window of both models to 5. This means that the range of the long-term dependency is greater than the context window of the model, requiring a long-term memory to solve. The models are trained for 100 batches on 5 different seeds. Learnt global positional embeddings are used in the X-Maze single-signal setting to minimise the differences between DT and DTi.

\subsubsection{Results and Discussion}

\begin{figure}[tp]
    \centering
    \includegraphics[width=1\linewidth]{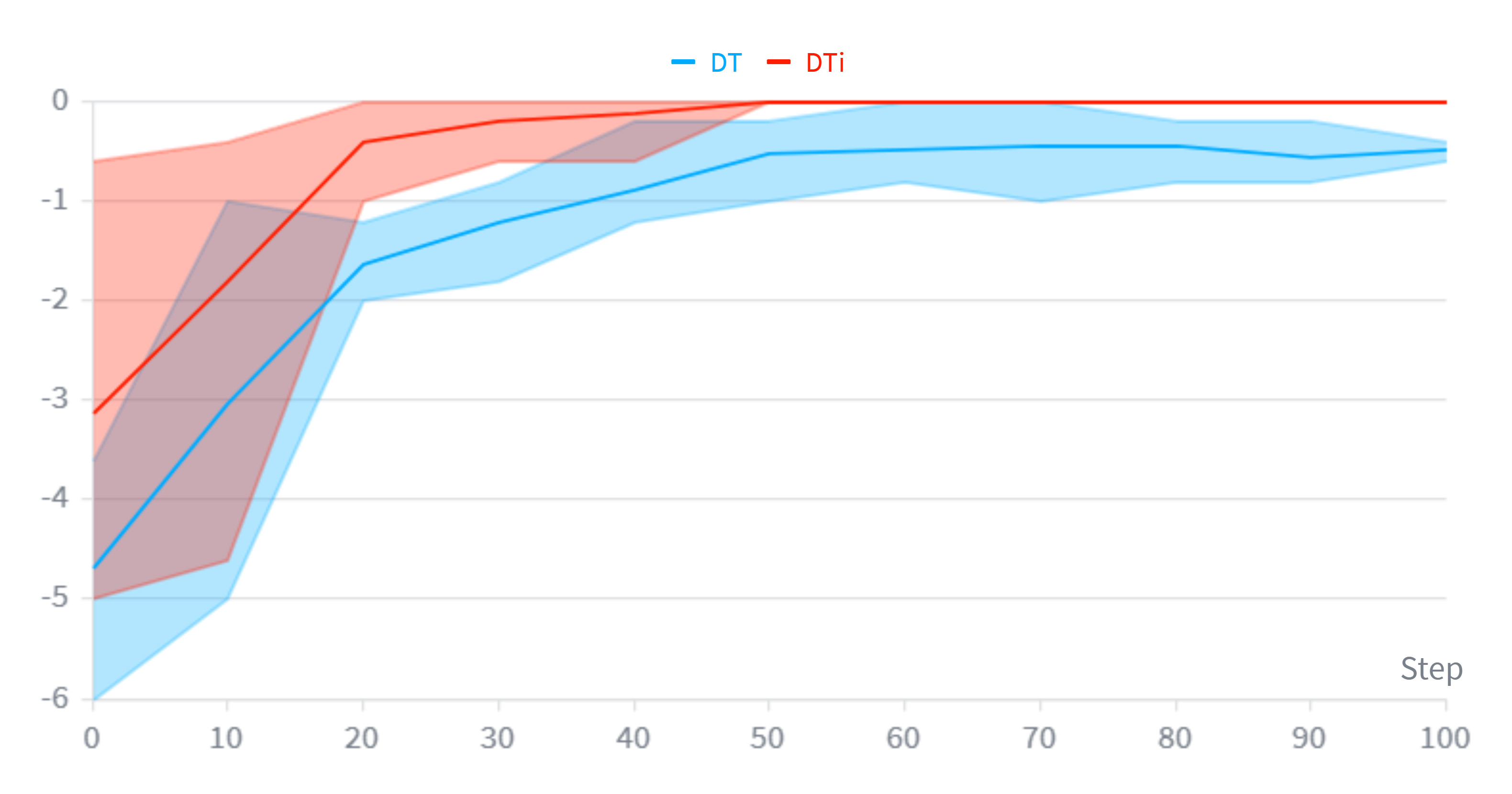}
    \caption{Mean episode return against training steps in T-Maze. The waiting phase length is 5, creating a dependency with a range of 6 timesteps. Tested models have a context length of 5, meaning they need a long-term memory in order to solve the environment. DTi solves the environment whilst DT does not, demonstrating that TTT memory works in RL.}
    \label{fig:xmaze_basic_mean_return}
\end{figure}

The mean return of 25 runs (5 per seed) over the course of training is shown in Figure \ref{fig:xmaze_basic_mean_return}. DT achieves a mean return of approximately -0.5, no better than random chance, whilst DTi achieve a mean return of 0---a result which would not be possible if the long-term memory module could not learn long-term dependencies. Although this is a very simple example, it clearly shows the advantage of DTi over DT and that the TTT framework had the capacity to handle long-term dependencies in RL environments.

\subsection{Control Tasks}

We test DTi in MuJoCo environments as a control. We do not expect the long-term memory mechanism to provide any benefit over standard DT \cite{DecisionTransformer} as these environments do not contain any long-term dependencies, however we run these tests as a sanity check to make sure that the memory mechanism does not have any unintended consequences.

\subsubsection{Setup}
We compare to DT as a base, as well as DTi with no TTT layers (i.e. just self-attention and persistent memory tokens) to separate the effects of the memory-mechanism and the other architecture choices of DTi.

We use chunked embeddings for DTi as the global sequence position should not be important in MuJoCo environments, where only the local-context is needed. For DT, we use learnt global timestep embeddings as per \citeauthor{DecisionTransformer} \cite{DecisionTransformer}. We use the same set of base parameters for all models to ensure a fair comparison.

More specifically, the environments tested are hopper-v0, halfcheetah-v0, and walkerd2D-v0 on both expert and medium level datasets from the Minari API \cite{minari}. We train for 40,001 batches and use an effective batch size of 2000 steps (100 context windows) for all models. For evaluation, we use the model weights from the last training iteration and report the average return from 5 episode on 5 different seeds, making for a total sample size of 25 episodes per environment.

\subsubsection{Results and Discussion}

The results, shown in Table \ref{tab:control_task_results}, show that there is no significant difference between the three models. This is not surprising as the MuJoCo environments contain no long-term dependencies, however there is an additional point here worth noting. As DTi processes chunks in sequences, the only context the first actions of each chunk gets is the last returns-to-go and observation, as well as what is stored in long-term memory. In the case where the memory is disabled, this causes a problem in partially observable environments as they rely on prior context.

This is not the case for MuJoCo environments as the full state is visible at each timestep, explaining why there is no decrease in return in the no-memory instance compared to standard DT. In conclusion, the long-term memory module does not significantly negatively impact behaviour that does not require memory, however the DTi architecture could potentially suffer in non-Markovian environments and likely does when used without memory.

\begin{table}[htb]
    \centering
    \caption{Mean Return on Control Tasks}
    \tiny
    \begin{tabular}{c c | c c c c}
        \toprule
        Env & Dataset & DT & DTi (no mem) & DTi \\
        \midrule
        \multirow{2}{*}{Hopper}
        & E & $3899 \pm 383$ & $3872 \pm 391$  & $3880 \pm 479 $ \\
        & M & $2248 \pm 862$ & $3379 \pm 491$ & $3355 \pm 511$ \\
        \multirow{2}{*}{Halfcheetah}
        & E & $7900 \pm 1099$ & $7072\pm152$ & $7028\pm 124.6$ \\
        & M & $7380 \pm 1647$ & $6713\pm207$ & $6608 \pm 134$ \\
        \multirow{2}{*}{Walker2D}
        & E & $6015.8 \pm 472.5$ & $5750\pm233$ & $5848\pm228$ \\
        & M & $5947\pm 73$ & $5899\pm 136$ & $5851 \pm 123.4$ \\
        \bottomrule
    \end{tabular}
    \label{tab:control_task_results}
\end{table}

\subsection{Additional Gate Visualisation Figures} \label{apx:experiments_gate_visualisation}
\begin{figure}[htb]
    \centering
    \includegraphics[width=1\linewidth]{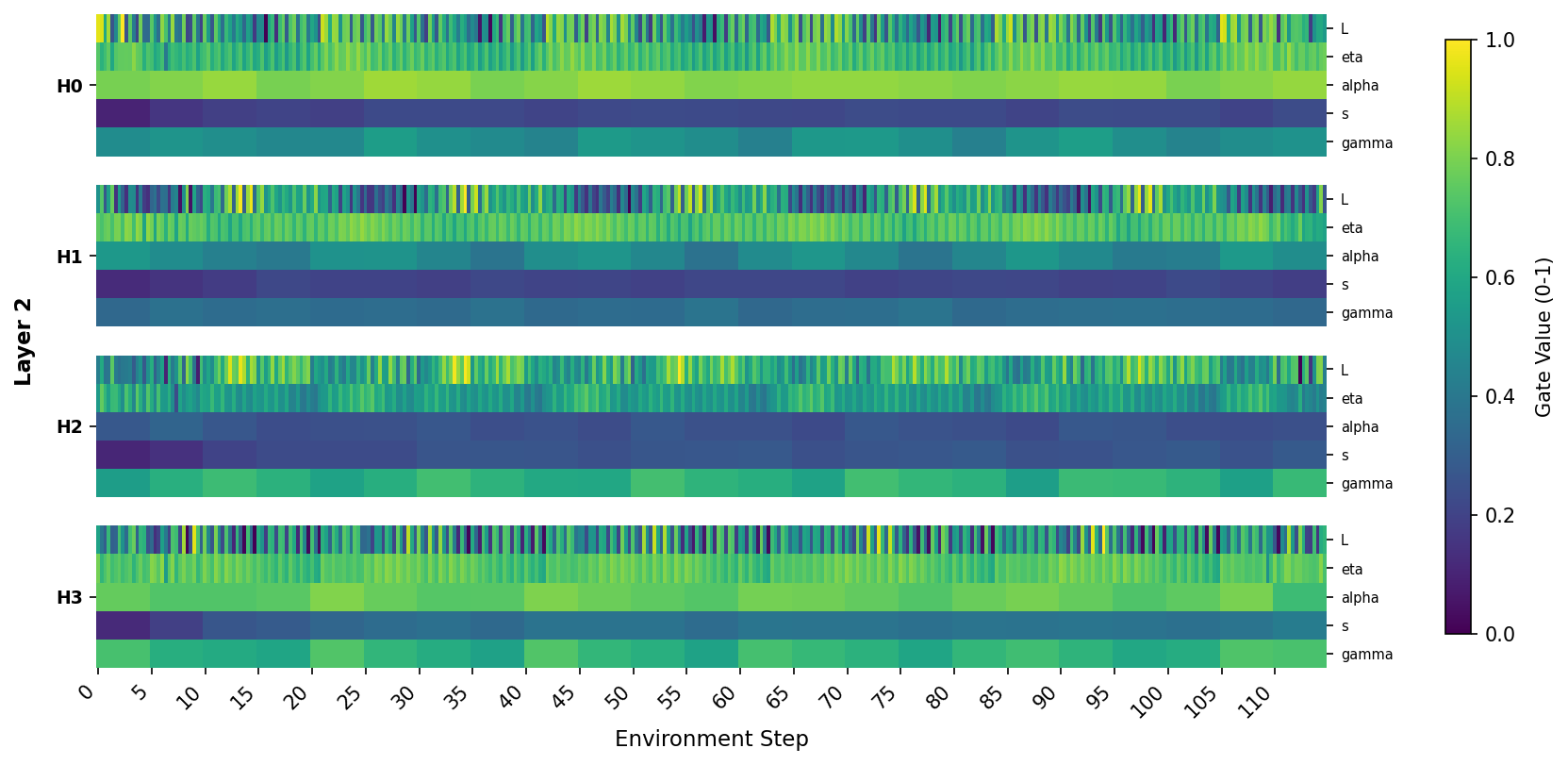}
    \caption{Visualisation of a failed run with momentum in X-Maze with a waiting phase of length 100. No head learns to maintain momentum or to minimise the the forget gate.}
    \label{fig:gates_100_failure}
\end{figure}

\begin{figure}[htb]
    \centering
    \includegraphics[width=1\linewidth]{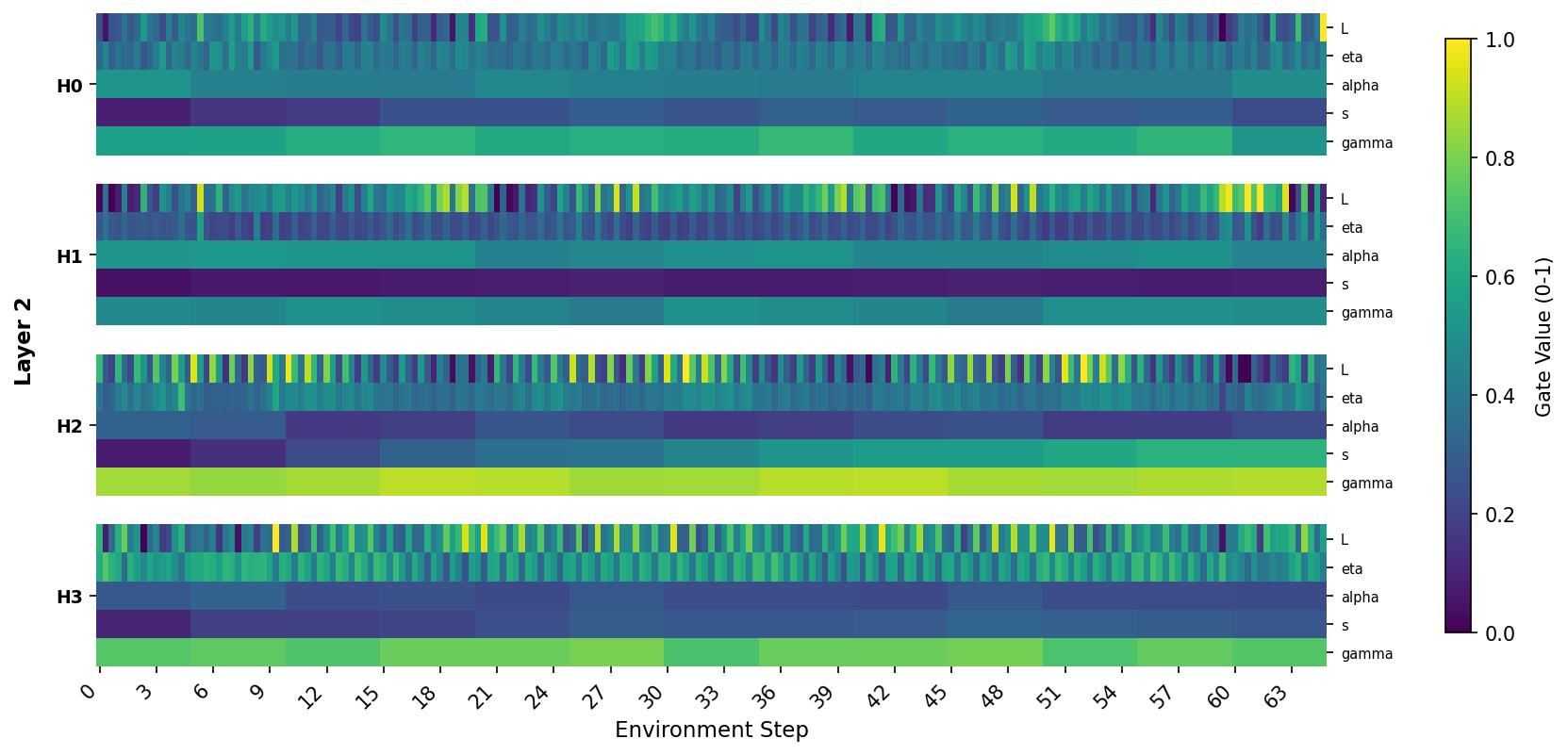}
    \caption{Visualisation of a successful run in X-Maze with a waiting phase of length 50. The same observations as the waiting phase length 100 case apply.}
    \label{fig:gates_50_success}
\end{figure}

\begin{figure}[htb]
    \centering
    \includegraphics[width=1\linewidth]{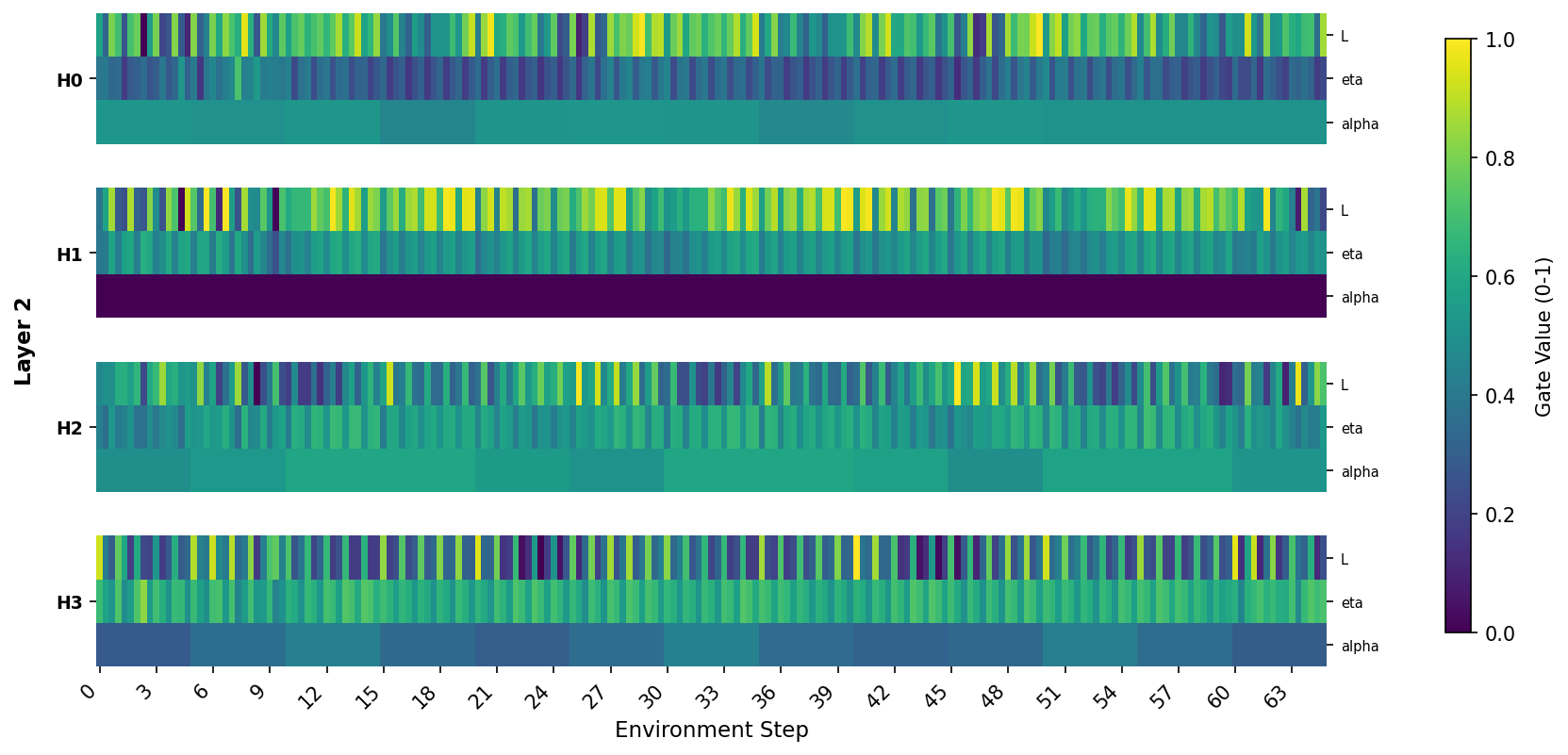}
    \caption{Visualisation of a successful run of DTi without momentum in X-Maze with a waiting phase of length 50. The update rule does not contain $s_t$ or $\gamma_t$. One head has learnt a small value for the forgetting gate.}
    \label{fig:gates_50_no_momentum}
\end{figure}

We provide additional figures relevant to the gate visualisation experiment in Section \ref{sec:experiments_gate_visualisation}, displaying some of the observations we describe. Figure \ref{fig:gates_100_failure} shows a failure case with momentum, Figure \ref{fig:gates_50_success} shows a success case with momentum in the 50 waiting phase length environment, and Figure \ref{fig:gates_50_no_momentum} shows a success case without momentum in the same environment. 

\end{document}